\documentclass[10pt,twocolumn,letterpaper]{article}

\usepackage{cvpr}              

\usepackage{graphicx}
\usepackage{amsmath}
\usepackage{amssymb}
\usepackage{booktabs}
\usepackage{multirow}
\usepackage{color}
\usepackage{arydshln}    
\usepackage[accsupp]{axessibility}  

\usepackage[pagebackref=true,breaklinks=true,colorlinks,bookmarks=false]{hyperref}
\usepackage[capitalize]{cleveref}
\crefname{section}{Sec.}{Secs.}
\Crefname{section}{Section}{Sections}
\Crefname{table}{Table}{Tables}
\crefname{table}{Tab.}{Tabs.}

\def\cvprPaperID{2026} 
\def\confName{CVPR}
\def\confYear{2023}

\begin{document}

\title{Progressive Semantic-Visual Mutual Adaption for Generalized Zero-Shot Learning}

\author{Man Liu$^{1,2}$, Feng Li$^{3}$, Chunjie Zhang$^{1,2}$, Yunchao Wei$^{1,2}$, Huihui Bai$^{1,2}$\thanks{Corresponding author}, Yao Zhao$^{1,2}$\\
	$^{1}$Institute of Information Science, Beijing Jiaotong University, Beijing, China\\
	$^{2}$Beijing Key Laboratory of Advanced Information Science and Network  Technology, Beijing,  China\\
	$^{3}$Hefei University of Technology, Hefei, China\\
	{\tt\small \{manliu, cjzhang, yunchao.wei, hhbai, yzhao\}@bjtu.edu.cn \quad fengli@hfut.edu.cn}
}
\maketitle

\begin{abstract}
Generalized Zero-Shot Learning (GZSL) identifies unseen categories by knowledge transferred from the seen domain, relying on the intrinsic interactions between visual and semantic information.
 Prior works mainly localize regions corresponding to the sharing attributes. 
 When various visual appearances correspond to the same attribute, the sharing attributes inevitably introduce semantic ambiguity, hampering the exploration of accurate semantic-visual interactions.
 In this paper, we deploy the dual semantic-visual transformer module (DSVTM) to progressively model the correspondences between attribute prototypes and visual features, constituting a progressive semantic-visual mutual adaption (PSVMA) network for semantic disambiguation and knowledge transferability improvement. 
Specifically, DSVTM devises an instance-motivated semantic encoder that learns instance-centric prototypes to adapt to different images, enabling the recast of the unmatched semantic-visual pair into the matched one. 
Then, a semantic-motivated instance decoder strengthens accurate cross-domain interactions between the matched pair for semantic-related instance adaption, encouraging the generation of unambiguous visual representations.
Moreover, to mitigate the bias towards seen classes in GZSL, a debiasing loss is proposed to pursue response consistency between seen and unseen predictions. The PSVMA consistently yields superior performances against other state-of-the-art methods.
Code will be available at: {\small\url{{https://github.com/ManLiuCoder/PSVMA.}}}
\end{abstract}

\section{Introduction}
\label{sec:intro}
\begin{figure}[t]
	\centering
	\includegraphics[width=1.0\linewidth]{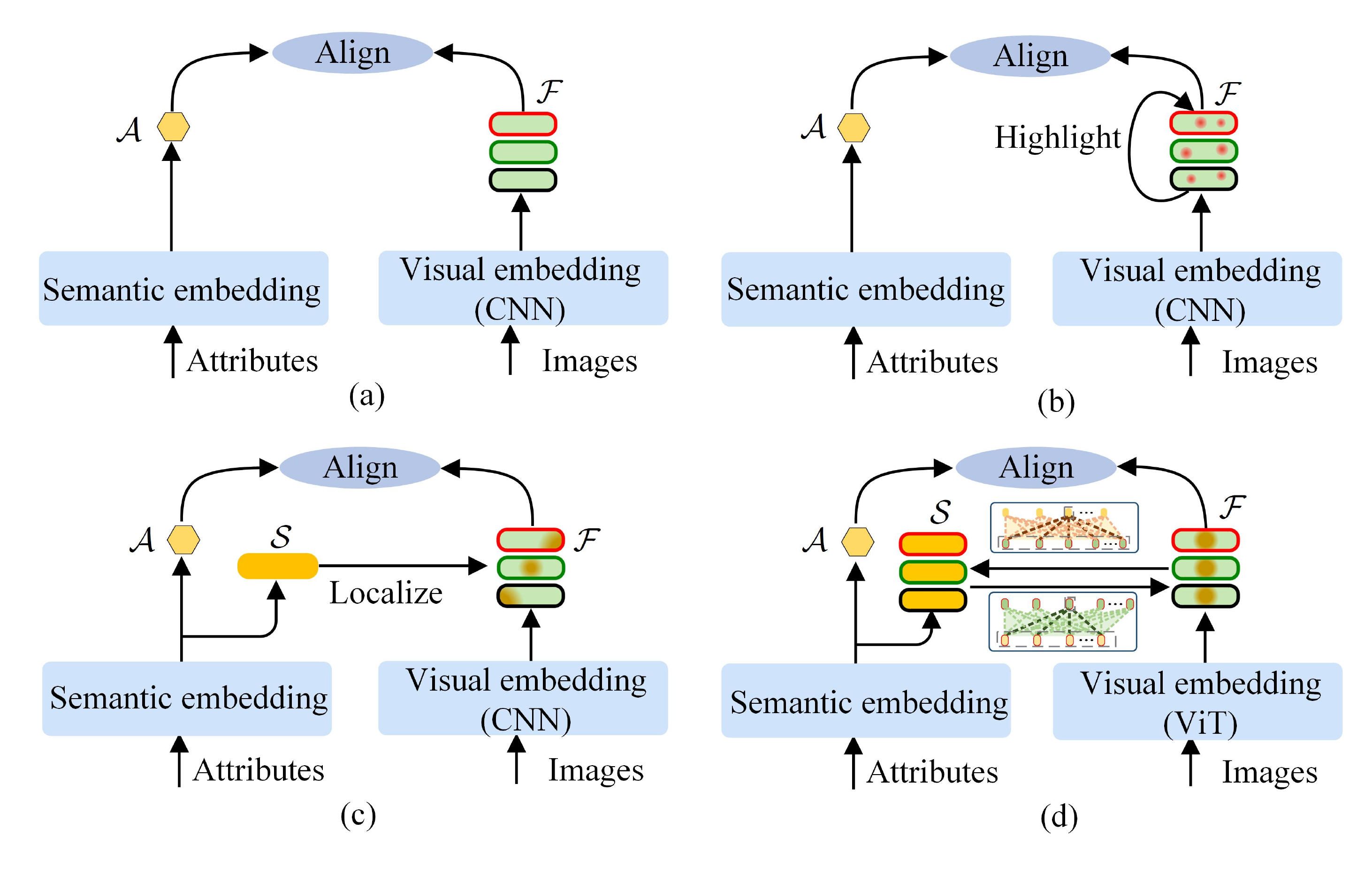}
	\vspace{-8mm}
	\caption{The embedding-based models for GZSL. (a) The early embedding-based method. (b) Part-based methods via attention mechanisms. (c) Semantic-guided methods. (d) Our PSVMA. $\mathcal{A}$, $S$, $F$ denote the category attribute prototypes, sharing attributes, and visual features, respectively.
		The PSVMA progressively performs semantic-visual mutual adaption for semantic disambiguation and knowledge transferability improvement.}\vspace{-7mm}
	\label{fig:motivation}
\end{figure}
Generalized Zero-Shot Learning (GZSL) \cite{palatucci2009zero} aims to recognize images belonging to both seen and unseen categories, solely relying on the seen domain data.
Freed from the requirement of enormous manually-labeled data, GZSL has extensively attracted increasing attention as a challenging recognition task that mimics human cognitive abilities \cite{lampert2009learning}.
As unseen images are not available during training,
knowledge transfer from the seen to unseen domains is achieved via auxiliary semantic information (\emph{i.e.}, category attributes \cite{lampert2009learning,farhadi2009describing}, text descriptions\cite{reed2016learning,lei2015predicting}, and word embedding \cite{socher2013zero,mikolov2013efficient,mikolov2013distributed}).

Early embedding-based methods \cite{akata2013label,SJE2015,xian2016latent,zhang2017learning} 
embed category attributes and visual images and learn to align global visual representations with corresponding category prototypes, as shown in \cref{fig:motivation} (a).
Nevertheless, 
the global information is insufficient to mine fine-grained discriminative features which are beneficial to capture the subtle discrepancies between seen and unseen classes.
To solve this issue, part-based learning strategies have been leveraged to explore distinct local features.
Some works \cite{LDF2018,SGMA2019,AREN2019,LFGAA2019,RGEN2020,DVBE2020} apply attention mechanisms to highlight distinctive areas, as shown in \cref{fig:motivation} (b).
These methods fail to develop the deep correspondence between visual and attribute features, which results in biased recognition of seen classes.
More recently, semantic-guided approaches (see \cref{fig:motivation} (c)) are proposed to employ the sharing attribute and localize specific attribute-related regions \cite{APN2020,xu2022attribute,DAZLE2020,GEM2021,MSDN2022}.
They establish interactions between the sharing attributes and visual features during localization, further narrowing the cross-domain gap. 
Actually, various visual appearances correspond to the same sharing attribute descriptor. For example, for the attribute descriptor ``tail'', the visual presentations of a dolphin's and rat's tail exhibit differently.
The above methods are suboptimal to build matched visual-semantic pairs and inclined to generate ambiguous semantic representations.
Further,
this semantic ambiguity can hamper cross-domain interactions based on unmatched visual-semantic pairs, which is detrimental to the knowledge transferring from seen to unseen classes.

To tackle this problem, we propose a progressive semantic-visual mutual adaption (PSVMA) network, as shown in \cref{fig:motivation} (d), to progressively adapt the sharing attributes and image features.
Specifically, inspired by the powerful ability of the vision transformers (ViT) \cite{ViT2020} to capture global dependencies,
we apply ViT for visual embedding and extract image patch features for the interaction with semantic attributes.
With the embedded visual and attribute features, we devise the dual semantic-visual transformer module (DSVTM) in PSVMA, which consists of an instance-motivated semantic encoder (IMSE) and a semantic-motivated instance decoder (SMID).

Concretely, in IMSE, we first perform instance-aware semantic attention to adapt the sharing attributes to various visual features. 
Based on the interrelationship between attribute groups, we further introduce attribute communication and activation to promote the compactness between attributes.
In this way, IMSE recurrently converts the sharing attributes into instance-centric semantic features and recasts the unmatched semantic-visual pair into the matched one, alleviating the problem of semantic ambiguity.
Subsequently, SMID explores the cross-domain correspondences between each visual patch and all matched attributes for semantic-related instance adaption, providing accurate semantic-visual interactions. 
Combined with the refinement of the patch mixing and activation in SMID, the visual representation is eventually adapted to be unambiguous and discriminative.
In addition, we design a novel debiasing loss for PSVMA to assist the process of knowledge transfer by pursuing the distribution consistency of inferred scores, mitigating the common bias towards seen domains.
Consequently, PSVMA can effectively achieve semantic disambiguation and improve knowledge transferability by progressive semantic-visual mutual adaption, gaining more accurate inferences for both seen and unseen categories.

Our key contributions can be summarized as follows: (1) We propose a progressive semantic-visual mutual adaption (PSVMA) network that deploys the dual semantic-visual transformer module (DSVTM) to alleviate semantic ambiguity and strengthen feature transferability through mutual adaption.
(2) The sharing attributes are converted into instance-centric attributes to adapt to different visual images, enabling the recast of the unmatched semantic-visual pair into the matched one. Furthermore, accurate cross-domain correspondence is constructed to acquire transferable and unambiguous visual features. 
(3) Extensive experiments over common benchmarks demonstrate the effectiveness of our PSVMA with superior performance. 
Particularly, our method achieves 75.4\% for the harmonic mean on the popular benchmark AwA2, outperforming previous competitive solutions by more than 2.3\%.

\begin{figure*}[t]
	\centering
	\includegraphics[width=0.9\linewidth]{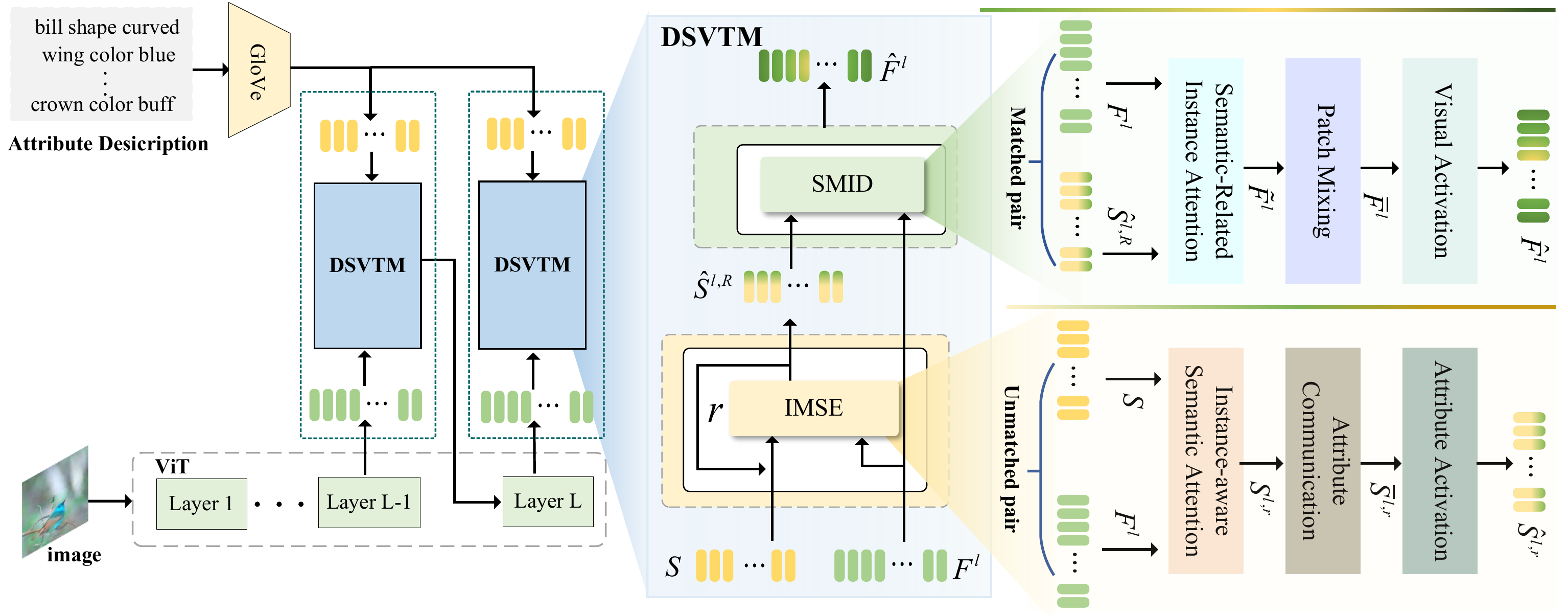}\vspace{-2mm}
	\caption{The framework of our proposed PSVMA. PSVMA deploys DSVTM between different visual layers and attribute prototypes, encouraging a progressive
		augmentation for semantic disambiguation and
		transferability improvement. The IMSE in DSVTM progressively learns the instance-centric semantics to acquire a matched semantic-visual pair. The SMID in DSVTM constructs accurate cross-domain interactions and learns unambiguous visual representations.}\vspace{-6mm}
	\label{fig:framework}
\end{figure*}

\section{Related work}
\label{sec:formatting}

\subsection{Generalized Zero-Shot Learning}
To transfer knowledge learned from the seen domain to the unseen domain, 
semantic information assumes a crucial role in providing a common space to describe seen and unseen categories.
With the category attribute prototypes, generative GZSL approaches synthesize visual features of extra unseen categories by generative adversarial nets \cite{Composer2020,LsrGAN2020,CEGZSL2021}, variational auto-encoders\cite{OCD2020,HSVA2021,SDGZSL2021}, or a combination of both \cite{TF-VAEGAN2020,FREE2021}. 
Although these methods compensate for the absence of the unseen domain during training, the introduction of extra data converts the GZSL problem into a fully supervised task.

The embedding-based method is the other mainstream branch for GZSL that projects and aligns information originating from visual and semantic domains.
Early works \cite{akata2013label,SJE2015,xian2016latent,zhang2017learning} directly map the global visual and semantic features into a common space for category predictions.
Global visual information, however, falls short in capturing subtle but substantial differences between categories, weakening discriminative representations.
To highlight discriminative visual regions, recent efforts have attempted part-based techniques.
Some works
\cite{LDF2018,SGMA2019} crop and zoom in on significant local areas employing coordinate positions obtained by attention mechanisms.
Distinctive visual features are also emphasized by graph networks \cite{RGEN2020,hu2021graph} or attention guidance \cite{AREN2019,LFGAA2019,DVBE2020}.
Furthermore, 
the sharing attribute prototypes, which are the same for all input images, have been introduced in semantic-guided methods \cite{APN2020,DPPN2021,xu2022attribute,DAZLE2020,GEM2021,MSDN2022} to localize attribute-related regions.
Among these methods, DPPN\cite{DPPN2021} updates attribute prototypes and achieves superior performance. 
However, DPPN ignores the deeply mutual interaction between semantic and visual domains, which limits the capability of the alleviation for semantic ambiguity.

\subsection{Transformers in GZSL}
Transformers \cite{vaswani2017attention} have a strong track record of success in Natural Language Processing (NLP) and have gradually imposed remarkable achievements in computer vision tasks \cite{ViT2020,cpcnn,dong2022incremental,gupta2022ow}.
Unlike CNNs, which are regarded as hierarchical ensembles of local features, transformers with cascaded architectures are encouraged to develop global-range relationships through the contribution of self-attention mechanisms.
Despite the effectiveness of the transformer's architecture (such as the vision transformer (ViT)\cite{ViT2020}), research on GZSL has lagged behind, with just a tiny amount of work \cite{alamri2021multi,alamri2021implicit,chen2022duet} using the ViT as a visual backbone. 
ViT-ZSL \cite{alamri2021multi} directly aligns the patch tokens of ViT with the attribute information and maps the global features of the classification token to the semantic space for category prediction.
IEAM-ZSL \cite{alamri2021implicit} not only captures the explicit attention by ViT, but also constructs another implicit attention to improve the recognition of unseen categories.
DUET \cite{chen2022duet} proposes a cross-modal mask reconstruction module to transfer knowledge from the semantic domain to the visual domain.
These works verify that, compared to CNNs, 
ViT specifically attends to image patches linked to category prototypes in GZSL.
However, they neglect the semantic ambiguity problem and fail to construct matched semantic-visual correspondences, limiting transferability and discriminability.
Additionally, the ViT model they applied is pre-trained on ImageNet-21k, which generates information leakage and leads to the incomplete GZSL problem.

\section{Methodology}
\noindent
\textbf{Problem Setting.}
GZSL attempts to identify unseen categories by the knowledge transferred from seen domain $\mathcal{D}^s$ to unseen domain $\mathcal{D}^u$. 
$\mathcal{D}^s=\{(x, y,a_y )|x \in \mathcal{X}^{s},y \in \mathcal{Y}^{s},a_y \in \mathcal{A}^{s}\}$, where $x$ refers to an image in $\mathcal{X}^{s} $, $y$ and $a_y$ refer to the corresponding label and category attributes.
Here, $\mathcal{D}^u=\{(x^{u}, u,a_{u} )\}$, $x^{u} \in \mathcal{X}^{u}$, $u \in \mathcal{Y}^{u}$, $a_{u} \in \mathcal{A}^{u}$, and $\mathcal{A}=\mathcal{A}^{s} \cup \mathcal{A}^{u}$.
Let $S$ denote the sharing attribute prototypes to describe the word vectors of each attribute, which are abstracted by a language model GloVe\cite{pennington2014glove}. 
In GZSL, the category space is disjoint between seen and unseen domain ($\mathcal{Y}^{s} \cap \mathcal{Y}^{u}=\varnothing$), while the testing data contains both seen and unseen categories ($\mathcal{Y}=\mathcal{Y}^{s} \cup \mathcal{Y}^{u}$). 
Therefore, an important problem is the seen-unseen bias, \emph{i.e.}, testing samples are more likely to be assigned to the seen categories observed during training.
The goal of this work is to design an effective framework that explores semantic-visual interactions for unbiased GZSL.

\noindent
\textbf{Overview.}
We first present the overall pipeline which is a progressive semantic-visual mutual adaption (PSVMA) network for GZSL (see \cref{fig:framework}).
PSVMA expects two inputs: visual features $F^l \in \mathbb{R}^{N_v \times D}$ and sharing attribute prototypes $S\in \mathbb{R}^{N_s \times D}$, which are obtained by a ViT \cite{ViT2020} visual backbone and GloVe\cite{pennington2014glove}, respectively.
Here, $N_v$ and $N_s$ denote the patch length and attribute prototypes with $D$ dimensional vectors, respectively.
Noted that $N_v$ does not contain a class token that is in vanilla ViT and $l$ refers to the $l$-th transformer layer in ViT.
With $F^l$ and $S$,
we devise the dual semantic-visual transformer module (DSVTM) to improve the visual-semantic alignment and discover discriminative attribute-related visual representations.
As shown in \cref{fig:framework}, DSVTM is a transformer-based structure that contains an instance-motivated semantic encoder (IMSE) and a semantic-motivated instance decoder (SMID), pursuing semantic and visual mutual adaption for the alleviation of semantic ambiguity. 
After progressive enhancements by DSVTMs, a classification head is applied for inferring.
\begin{figure}[t]
	\centering
	\includegraphics[width=0.9\linewidth]{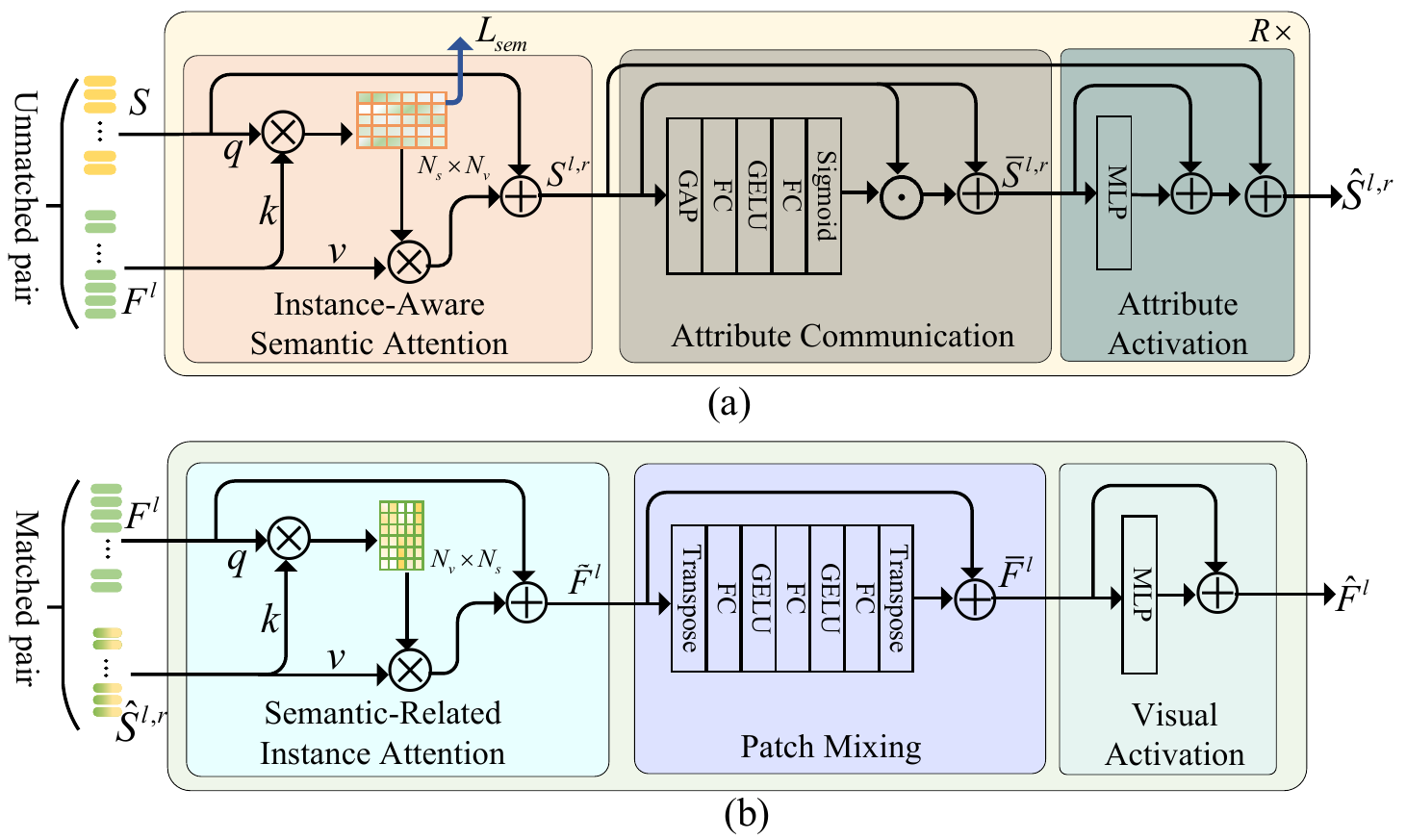}\vspace{-2mm}
	\caption{The architectures of (a) IMSE and (b) SMID.} \vspace{-5mm}
	\label{fig:en_de}
\end{figure}
\subsection{Instance-Motivated Semantic Encoder}
In DSVTM, the proposed IMSE aims at progressively learning instance-centric prototypes to produce accurately matched semantic-visual pairs in a recurrent manner with $r (r=1,..., R)$ loops.
As shown in \cref{fig:en_de} (a),
IMSE contains the instance-aware semantic attention, attribute communication and activation, which are elaborated as follows.

\noindent
\textbf{Instance-Aware Semantic Attention.} 
To adapt the sharing attributes $S$ to different instance features $F^l$, IMSE first executes cross-attention to learn attentive semantic representations based on instance features.
For the $i$-th attribute $S_i$, we search for the most relevant patch $F^l_j$ by modeling the relevance $M^{l,r}_{(i,j)}$: 
\begin{equation}
M^{l,r}_{(i,j)}=q({\rm LN}(S_i))\cdot k({\rm LN}(F^l_j))^T
\label{eq:M}
\end{equation}  
where LN denotes the Layer Normalization, $T$ denotes the transpose function, $q(\cdot)$ and $k(\cdot)$ are the linear mapping functions for the query and key. 
$M^{l,r}_{(i,j)}$ indicates the localized region $F^l_{j}$ related to the attribute descriptor $S_{i}$, forming an affinity matrix
$M^{l,r}\in \mathbb{R}^{N_s \times N_v}$.
To encourage the localization ability of patches related to attributes, 
we apply a semantic alignment loss to align $M^{l,r}$ with its category prototypes $a_{y}$:
\begin{equation}
\mathcal{L}^{l,r}_{sem}=\|\hbar(M^{l,r})-a_y\|_{2}^{2}
\label{eq:L_sem}
\end{equation} 
where $\hbar$ is the 1-dimensional global max pooling (GMP) operation.
Then, $M^{l,r}$ is applied to select distinct visual patches in $F^l$ related to each attribute,
packed together into instance-related attribute prototypes $S^{l,r} \in \mathbb{R}^{N_s \times D}$ with a residual connection:
\begin{equation}
S^{l,r}={\rm softmax}(M^{l,r})\cdot v({\rm LN}(F^l)) + S
\label{eq:2}
\end{equation}
where $v(\cdot)$ is the linear mapping functions for the value.
Compared to the original sharing attribute prototype $S$, 
the instance-motivated semantic attribute $S^{l,r}$ is more discriminative and more closely linked to specific instance.

\noindent
\textbf{Attribute Communication and Activation.}
As attribute descriptors are interdependent, IMSE then conducts attribute communication to compact the relevant attributes and scatter the irrelevant attributes via a group compact attention $f_{gc}(\cdot)$:
\begin{align}
f_{gc}({S}^{l,r})&= {\rm sigmoid}(\sigma (\hbar({S}^{l,r}) \cdot W_{p1})\cdot W_{p2})
\label{eq:f_s} \\
\bar{S}^{l,r}&=f_{gc}({S}^{l,r})\cdot{S}^{l,r}+{S}^{l,r}
\label{eq:S}
\end{align}
where $\sigma$ is the GELU \cite{gelu} function. $W_{p1} \in \mathbb{R}^{N_S \times \frac{N_S}{N_g}}$ and $W_{p2} \in \mathbb{R}^{\frac{N_S}{N_g} \times N_S}$ are the parameters of two fully-connected (FC) layers, respectively. $N_g$ denotes the number of attribute groups given in the datasets (\emph{e.g.}, 28 groups for 312 attributes on CUB dataset \cite{DatasetCUB}).
To make use of the compacted attribute prototypes $\bar{S}^{l,r}$, we further activate significant features and squeeze out trivial ones in each attribute by an MLP layer: 
\begin{equation}
\hat{S}^{l,r}={\rm MLP}(\bar{S}^{l,r})+ \bar{S}^{l,r} + {S}^{l,r}
\label{eq:mlp}
\end{equation} 
Here, cooperated with residual connections of $\bar{S}^{l,r}$ and ${S}^{l,r}$, more instance-aware information can be preserved.

By implementing IMSE recurrently, we can progressively adapt the sharing attributes by observing previously adapted ones and visual instances,
distilling instance-centric attribute prototypes. 
With the produced $\hat{S}^{l,R}$, the unmatched semantic-visual pairs $(S, F^l)$ can be recast into matched pairs $(\hat{S}^{l,R}, F^l)$ ultimately.

\subsection{Semantic-Motivated Instance Decoder}
Given the matched semantic-visual pair $(\hat{S}^{l,R}, F^l)$ from IMSE, 
we design a SMID to strengthen cross-domain interactions and learn unambiguous visual representations via the semantic-related instance attention, patch mixing and activation, as shown in \cref{fig:en_de} (b).

\noindent
\textbf{Semantic-Related Instance Attention.}
To acquire semantic-related visual representations, 
we first model the cross-domain correspondence between the matched semantic-visual pair $(\hat{S}^{l,R}, F^l)$ via a cross-attention. 
Compared to the attention in IMSE (\cref{eq:M}), here, we focus on instance-centric attributes with respect to each visual patch and obtain attention weights $\bar{M}^{l,R}$:
\begin{equation}
{\bar{M}^{l,R}}=q({\rm LN}(F^l))\cdot k({\rm LN}(\hat{S}^{l,R}))^T
\label{eq:M'}
\end{equation} 
$\bar{M}^{l,R}$ is applied to select information in $\hat{S}^{l,R}$ and helps to aggregate significant semantic characteristics into the visual patches:
\begin{equation}
\tilde{F^l}={\rm softmax}(\bar{M}^{l,R})\cdot v({\rm LN}(\hat{S}^{l,R}) +F^l
\label{eq:F1}
\end{equation} 
where $\tilde{F^l}$ denotes the visual instance representation that is aligned with the learned instance-centric attributes $\hat{S}^{l,R}$.
By such semantic-related attention, we can construct more accurate visual-semantic interactions and gather powerful matching attribute information in $\tilde{F}^l$.  

\noindent
\textbf{Patch Mixing and Activation.} 
Considering that the detailed information between different patches is essential for fine-grained recognition, 
we propose a patch mixing and activation module to expand and refine the association between patches.
Inspired by the concept of the manifold of interest in \cite{mobilenetv2}, we mix patches by an inverted residual layer with a linear bottleneck to improve the representation power.
This process can be formulated as:
\begin{align}
		F^l_e &= f_{e}(({\tilde{F}^l})^T)=\sigma(({\tilde{F}^l})^T\cdot W_e)\\
		F^l_s &=f_{s}(F^l_e)=\sigma(F^l_e\cdot W_s)\\
		F^l_n &=f_{n}(F^l_s)=F^l_s\cdot W_n
	\end{align}
where $f_e(\cdot)$ is an expansion layer consisting of an FC layer with parameters $W_{e} \in \mathbb{R}^{N_v \times N_{h}}$ followed by an activation function.
Thus, the length of patch is expanded to a higher dimension $N_{h}(N_{h}>N_v)$ for the subsequent information filtering implemented by a selection layer $f_s(\cdot)$.
 Then, the mixed and selected patches are projected back to the original low-dimension $N_v$ by a narrow linear bottleneck $f_n(\cdot)$. 
We utilize a shortcut to
preserve complete information and produce $\bar{F}^l = (F^l_n)^T + \tilde F^l$. 
After that,
the refined features in each visual patch are activated by an MLP layer with a residual connection:
\begin{equation}
\hat{F^l}={\rm MLP}(\bar{F}^l)+ \bar{F}^l
\label{eq:hat_F}
\end{equation}

With SMID, we can take effects on the visual instance based on adapted semantic attributes along the spatial dimension, realizing the attentive visual features to keep with the attribute information.
By unifying IMSE and SMID in $Z$ cascaded DSVTMs, our network can achieve progressive semantic-visual mutual adaption to generate unambiguous and transferable visual representations.

\noindent
\textbf{Classification Head.} 
As shown in \cref{fig:framework},
after progressive learning of $Z$ DSVTM in visual layers and receiving the final visual representation by the last DSVTM (denoted as $\hat{F}^L$), a classification head $f_c(\cdot)$ conducted on $\hat{F}^L$ is adopted for instance category prediction. 
\begin{equation}
f_c(\hat{F}^L)=\hbar((\hat{F}^L)^T)W
\label{eq:f_c}
\end{equation}
where $W$ denotes the learnable parameter with the size of $D \times N_s$, which is applied to project visual features into class embedding space. $f_c(\hat{F}^L)$ is the predicted probability of category attributes.
Then, we measure the cosine similarity $cos(\cdot)$ between $f_c(\hat{F}^L)$ and category prototypes $\mathcal{A}$ for classification:
\begin{equation}
score(\hat{y}|x) = \tau \cdot cos (f_c(\hat{F}^L), \mathcal{A}) 
\end{equation} 
where $\tau$ is the scaling factor. The category of instance $x$ is supervised by the classification loss ${\cal L}_{cls}$ defined as:
\begin{equation}
{{\cal L}_{cls}} = - \log \frac{{\exp \left( {score(y|x)} \right)}}{{\sum\limits_{\hat y \in {{\cal Y}^S}} {\exp } \left( {score(\hat y|x)} \right)}}
\end{equation}

\subsection{Model Optimization and Inference}
\noindent
\textbf{Optimization.} 
In addition to the semantic alignment loss and classification loss mentioned above, we design a debiasing loss $\mathcal{L}_{deb}$ to mitigate the seen-unseen bias.
To better balance the score dependency in the seen-unseen domain, $\mathcal{L}_{deb}$ is proposed to pursue the distribution consistency in terms of mean and variance:
\begin{equation}
{{\cal L}_{deb}}{\rm{ }} = \|{\alpha _s} - {\alpha _u}\|_2^2 + \|{\beta _s} - {\beta _u}\|_2^2
\end{equation}
$\alpha _s$ and $\beta _s$ denote the mean and variance value of seen predictions $score(\hat{y_s}|x, \hat{y_s} \in \mathcal{Y}^{s})$. $\alpha _u$ and $\beta _u$ denote the mean and variance value of $score(\hat{y_u}|x, \hat{y_u} \in \mathcal{Y}^{u})$.

Finally, the overall optimization goal can be defined as:
\begin{equation}
\mathcal{L}=\mathcal{L}_{cls} +\lambda _{sem}\mathcal{L}_{sem}+\lambda _{deb}\mathcal{L}_{deb}
\label{eq:loss}
\end{equation}
where $\lambda _{sem}$ and $\lambda _{deb}$ are the hyper-parameters for the semantic alignment loss $\mathcal{L}_{sem} = \sum\limits_{l = L-Z+1}^L {\sum\limits_{r = 1}^R {\mathcal{L}_{{sem}}^{l,r}} }$ and debiasing loss $\mathcal{L}_{deb}$.

\noindent
\textbf{Inference.} 
During training, the model merely learns about the knowledge of seen categories, whereas both seen and unseen categories are contained at inference time. 
Therefore, calibrated stacking (CS) \cite{2016cs} is applied to jointly define the category:
\begin{equation}
\tilde{y} =\underset{\hat{y} \in \mathcal{Y}^{s} \cup \mathcal{Y}^{u}}{\arg \max } (score(\hat{y}|x)-\gamma \mathbb{I}_{\mathcal{Y}^{S}}(\hat{y}))
\label{eq:infer}
\end{equation}
$\mathbb{I}_{\mathcal{Y}^{S}}(\cdot)$ denotes an indicator function, whose result is 1 when $\hat{y} \in \mathcal{Y^S}$ and 0 otherwise.
A calibrated factor $\gamma$ is applied to trade-off the calibration degree on seen categories and decides the category $\tilde{y}$ of a sample $x$. 

\begin{table}[]
	\centering
	\caption{Proposed Split (PS) of GZSL datasets to evaluate our network. $N_S$ and $N_g$ denote the number of attribute dimensions and attribute groups, respectively. $s$ and $u$ are the number of seen and unseen classes.}\vspace{-3mm}
	\renewcommand{\arraystretch}{1.0}
	\begin{tabular}{c|c|c|c}
		\hline
		Datasets      & classes ($s$ $|$ $u$) & images & $N_S$ ($N_g$) \\ \hline \hline
		CUB \cite{DatasetCUB}  & 200 (150 $|$ 50)     & 11,788     & 312 (28) \\
		SUN \cite{DatasetSUN}   & 717 (645 $|$ 72)     & 14,340   & 102 (4)    \\
		AwA2 \cite{DatasetAWA2}   & 50 (40 $|$ 10)      & 37,322     & 85 (9)    \\  \hline
	\end{tabular}\vspace{-5mm}
	\label{Tab:PS}
\end{table}

\section{Experiment}
\begin{table*}[t]
	\centering
	\renewcommand{\arraystretch}{1.2}
	\caption{ Experimental Results (\%) on public benchmarks. The best and second-best results are marked in \textcolor[rgb]{1,0,0}{red} and \textcolor[rgb]{0,0,1}{blue}, respectively. Methods belonging to generative and embedding-based frameworks (denoted as ``GEN.'' and ``EMB.'') are compared separately. $^\diamondsuit$ denotes the model is pre-trained on ImageNet-21k. $*$ indicates 2048 dimensional top-layer pooling units of ResNet101 without fine-tuning.}\vspace{-3mm}
	\label{Tab:results}
	\resizebox{\textwidth}{45mm}{
		\begin{tabular}{l|l|l|l|ccc|ccc|ccc}
			\hline
			& \multirow{2}{*}{Methods}                                  & \multirow{2}{*}{Backbone} & \multirow{2}{*}{Image size} & \multicolumn{3}{c|}{CUB}                                                  & \multicolumn{3}{c|}{SUN}                                                     & \multicolumn{3}{c}{AwA2}                                                                               \\ \cline{5-13} 
			&                                                           &                           &                             & $U$                                 & $S$                              & $H$    & $U$                                 & $S$                                 & $H$    & $U$                                 & $S$                                 & $H$                              \\ \hline \hline
			\multirow{5}{*}{\rotatebox{270}{GEN.}}   
			& LsrGAN (ECCV'20)  \cite{LsrGAN2020}      & ResNet101  & $*$  & 48.1   & 59.1   & 53.0 
			& 44.8   & 37.7   & 40.9 & -     & -     & -                              \\
			& CE-GZSL (CVPR'21) \cite{CEGZSL2021}      & ResNet101                 & $*$        & 63.9                              & 66.8                           & 65.3 & 48.8                              & 38.6                              & 43.1 & 63.1                              & 78.6                              & 70.0                           \\
			& FREE (ICCV'21) \cite{FREE2021}           & ResNet101                 & $*$             & 55.7                              & 59.9                           & 57.7 & 47.4                              & 37.2                              & 41.7 & 60.4                              & 75.4                              & 67.1                           \\
			& HSVA (NeurIPS'21) \cite{HSVA2021}        & ResNet101                 & $*$              & 52.7                              & 58.3                           & 55.3 & 48.6                              & 39.0 & 43.3 & 56.7                              & 79.8                              & 66.3                           \\
			& ICCE (CVPR'22) \cite{En-Compactness2022}        & ResNet101                 & $*$              & 67.3                              & 65.5                          &66.4 & -                              & - & - & 65.3                              & 82.3                              & 72.8                           \\ \hline
			\multirow{12}{*}{\rotatebox{270}{EMB.}} 
			& AREN (CVPR'19)  \cite{AREN2019}          & ResNet101                 & 224$\times$224                     & 63.2                              & 69.0                           & 66.0 & 40.3                              & 32.3                              & 35.9 & 54.7                              & 79.1                              & 64.7                           \\
			& DVBE (CVPR'20) \cite{DVBE2020}           & ResNet101                 & 448$\times$448                     & 53.2                              & 60.2                           & 56.5 & 45.0                              & 37.2                              & 40.7 & 63.6                              & 70.8                              & 67.0                           \\
			& DAZLE (CVPR'20) \cite{DAZLE2020}         & ResNet101                 & 224$\times$224                     & 56.7                              & 59.6                           & 58.1 &52.3 & 24.3                              & 33.2 & 60.3                              & 75.7                              & 67.1                           \\
			& APN (NeurIPS'20) \cite{APN2020}          & ResNet101                 & 224$\times$224                 & 65.3                              & 69.3                           & 67.2 & 41.9                              & 34.0                              & 37.6 & 56.5                              & 78.0                              & 65.5                           \\
			& GEM-ZSL (CVPR'21) \cite{GEM2021}         & ResNet101                 & 448$\times$448                  & 64.8                              & \textcolor[rgb]{0,0,1}{77.1}                           & 70.4 & 38.1                              & 35.7                              & 36.9 & 64.8 & 77.5                              & 70.6                           \\
			& DPPN (NeurIPS'21)  \cite{DPPN2021}       & ResNet101                 & 448$\times$448                     & \textcolor[rgb]{1,0,0}{70.2} & \textcolor[rgb]{0,0,1}{77.1}                           & \textcolor[rgb]{0,0,1}{73.5} & 47.9                              & 35.8                              & 41.0 & 63.1                              & 86.8                              & \textcolor[rgb]{0,0,1}{73.1} \\
			& TransZero (AAAI'22) \cite{TransZero2022} & ResNet101                 & 448$\times$448                 & 69.3                              & 68.3                           & 68.8 & \textcolor[rgb]{0,0,1}{52.6}    & 33.4                              & 40.8 & 61.3                              & 82.3                              & 70.2                           \\
			& MSDN (CVPR'22) \cite{MSDN2022}           & ResNet101                 & 448$\times$448               & 68.7                              & 67.5                           & 68.1 & 52.2                              & 34.2                              & 41.3 & 62.0                              & 74.5                              & 67.7                           \\
			& ViT-ZSL (IMVIP'21) \cite{alamri2021multi}           & ViT-Large$^\diamondsuit$                  & 224$\times$224               & 67.3                              & 75.2                           & 71.0 & 44.5                              & \textcolor[rgb]{1,0,0}{55.3}                              & 49.3 & 51.9                              & \textcolor[rgb]{1,0,0}{90.0}                              & 68.5                           \\
			& IEAM-ZSL (DGAM'21) \cite{alamri2021implicit}           & ViT-Large$^\diamondsuit$                  & 224$\times$224                & 68.6                              & 73.8                           & 71.1 & 48.2                              & \textcolor[rgb]{0,0,1}{54.7}                              & \textcolor[rgb]{0,0,1}{51.3} & 53.7                              & \textcolor[rgb]{0,0,1}{89.9}                              & 67.2                           \\
			& DUET (AAAI'23) \cite{chen2022duet} & ViT-Base$^\diamondsuit$                 & 224$\times$224               & 62.9                              & 72.8                           & 67.5 & 45.7                              & 45.8                              & 45.8 & 63.7                              & 84.7                              & 72.7                          \\
			& PSVMA (Ours)   & ViT-Base                       & 224$\times$224                     & \textcolor[rgb]{0,0,1}{70.1}                                  & \textcolor[rgb]{1,0,0}{77.8}                               & \textcolor[rgb]{1,0,0}{73.8}     &\textcolor[rgb]{1,0,0}{61.7}                                   &45.3                                   & \textcolor[rgb]{1,0,0}{52.3}     &\textcolor[rgb]{1,0,0}{73.6}                                   &77.3                                   & \textcolor[rgb]{1,0,0}{75.4}                               \\ \hline
	\end{tabular}}\vspace{-5mm}
\end{table*}
\subsection{Experimental Setup}
\noindent
\textbf{Datasets.}
We evaluate PSVMA on three benchmark datasets, \emph{i.e.}, Caltech-USCD Birds-200-2011 (CUB) \cite{DatasetCUB}, SUN Attribute (SUN) \cite{DatasetSUN}, Animals with Attributes2 (AwA2) \cite{DatasetAWA2}.
The seen-unseen classes division is set according to Proposed Split (PS) \cite{DatasetAWA2} as shown in \cref{Tab:PS}.

\noindent
\textbf{Metrics.}
Following \cite{DatasetAWA2}, we apply the harmonic mean (defined as $H=2 \times S \times U /(S+U)$)
to evaluate the performance of our framework under GZSL scenarios. $S$ and $U$ denote the Top-1 accuracy of seen and unseen classes, respectively.

\noindent
\textbf{Implementation Details.}
Unlike previous GZSL works that utilize ResNet\cite{Resnet2016} models as visual backbones,
we take ViT-Base \cite{ViT2020} model pre-trained on ImageNet-1k as the visual feature extractor. Note that, we discard the ViT model pre-trained on a large dataset, \emph{e.g.,} ImageNet-21k, where some classes overlap with unseen classes defined in \cite{DatasetAWA2}, leading to incomplete GZSL.
Our framework is
implemented with Pytorch over an Nvidia GeForce RTX 3090 GPU.
The factor $\gamma$ and $\tau$ are set following \cite{GEM2021}.

\subsection{Comparison with State-of-the-Arts}
\noindent
\textbf{Comparisons with CNN Backbones.}
Here, we compare our method with recent CNN-based methods which adopt ResNet101 as the backbone.
As shown in \cref{Tab:results},
our PSVMA achieves the best harmonic mean $H$ of 73.8\%, 52.3\% and 75.4\% on CUB, SUN and AwA2, respectively.
These results demonstrate the effectiveness of PSVMA for GZSL.
Moreover,
compared to the methods (\emph{e.g.,} APN\cite{APN2020}, GEM-ZSL\cite{GEM2021}, DPPN\cite{DPPN2021}, MSDN\cite{MSDN2022}, TransZero\cite{TransZero2022}) which utilize the sharing attribute prototypes, PSVMA obtains significant $H$ gains over 0.3\%, 11.0\%, and 2.3\% on CUB, SUN, and AwA2, respectively. 
This demonstrates that PSVMA can learn better instance-centric attributes for more accurate semantic-visual interactions, thus improving knowledge transferability.
Especially, even using the input image size of $224 \times 224$, our method achieves comparable performance to the most SOTA method DPPN \cite{DPPN2021} ($448 \times 448$) on CUB dataset and the best accuracy on other two datasets for unseen classes. 

\noindent
\textbf{Comparisons with ViT Backbones.}
To further investigate the superiority of our method, we also compare PSVMA with some ViT-based methods \cite{alamri2021multi,alamri2021implicit,chen2022duet}. 
Generally, PSVMA performs the best $U$ and $H$ on all datasets. 
We can see that the seen-unseen performance is not always consistent. Enhancing model transferability (increased $U$) may reduce discrimination (decreased $S$). This is because GZSL methods align with category attributes, but attribute labels of various categories are non-orthogonal to each other. Hence, we pursue a trade-off between the seen and unseen domains to improve overall $H$.
Besides, compared to ViT-ZSL \cite{alamri2021multi} and IEAM-ZSL \cite{alamri2021implicit} which apply a large ViT architecture (\emph{i.e.}, ViT-Large), PSVMA exceeds them by a significant margin.
Although IEAM-ZSL is carefully designed to improve the recognition of unseen categories by a self-supervised task, it shows lower performance than PSVMA with $U$ falling 1.5\%, 13.5\% and 19.9\% on CUB, SUN, and AwA2 datasets.
Noted that these compared ViT-based methods use the pre-trained models on ImageNet-21k, while our method only applies the backbone pre-trained on ImageNet-1k.

\begin{figure*}[t]
	\centering
	\includegraphics[width=0.9\linewidth]{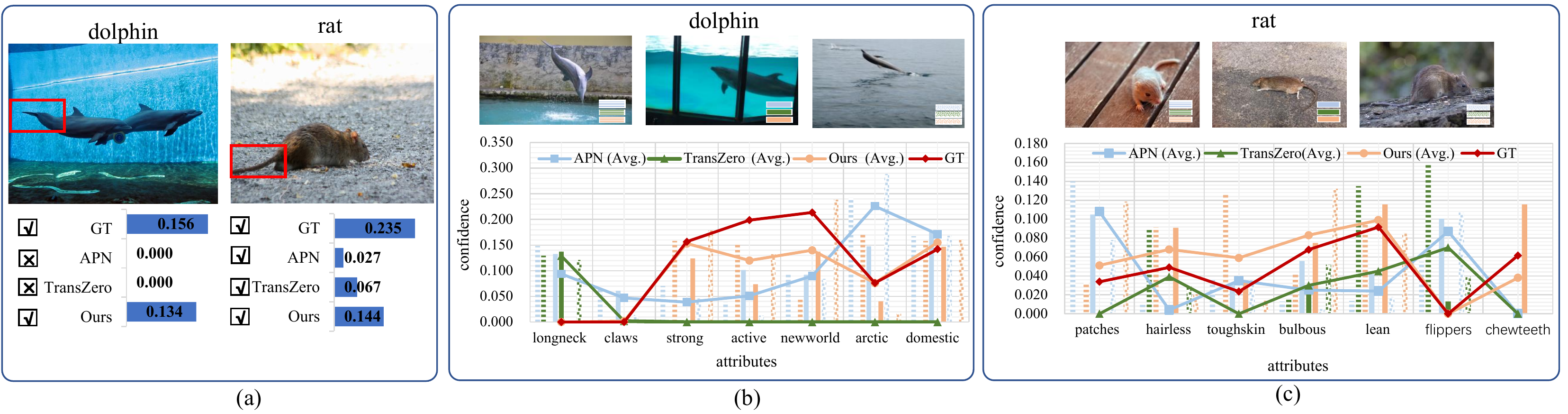}\vspace{-3mm}
	
	\caption{Visualization for attribute disambiguation. 
		(a) Inter-class disambiguation between the dolphin and rat for attribute ``tail''. Values in the bar represent the confidence of corresponding attributes.
		(b) and (c) show the Intra-class disambiguation in the dolphin and rat, receptively.
		The bar with a different mark corresponds to the image with the same mark. The blue, green and orange bars denote the results of APN, TransZero and ours, receptively. The line graph represents the average results of three randomly selected images.}\vspace{-5mm}
	\label{fig:amb}
\end{figure*}
 \subsection{Analysis of Semantic Disambiguation}

To intuitively provide the semantic disambiguation ability of our method, we calculate predicted probability of category attribute (\cref{eq:f_c}) as the confidence and compare with several methods including APN\cite{APN2020}, TransZero\cite{TransZero2022}. These two methods both use sharing attributes and have the same attribute prediction and category decision formulas.
As shown in \cref{fig:amb} (a), the attribute ``tail'' shows different appearances in a dolphin's and a rat's image (\textcolor[rgb]{1,0,0}{red box}). 
APN and TransZero fail to infer the ``tail'' in the dolphin, while our method predicts the attribute in both of the dolphin and rat correctly with closer confidence to GT (ground truth). 
 
Visual discrepancies for the same attribute information occur not only between classes but also within a class, especially for non-rigid objects with variable postures. Taking the dolphin as an example, \cref{fig:amb} (b) gives some attribute predictions of three randomly selected dolphin images. In the three intra-class instances, our method successfully determines that the dolphins do not have ``longneck'' properties yet all have a strong probability of being ``active'',``strong'', and ``new world''. 
Overall, the average attribute predictions of the three images are more consistent with the GT compared to APN and TransZero methods.
The similar intra-class disambiguation phenomenon can be observed in rats (see \cref{fig:amb} (c)). 
These demonstrate that the semantic-visual interactions explored by our matched semantic-visual pair are beneficial for the knowledge transferring process, encouraging inter-class and intra-class attribute disambiguation.
This verifies that our method can effectively alleviate the semantic ambiguity and achieve more accurate attribute prediction and category inference.

\subsection{Ablation Study}
\begin{table*}[]
	\centering
	\renewcommand{\arraystretch}{1.0}
	\caption{Analysis of each component in PSVMA. IASA and ACA denote the instance-aware semantic attention, and the attribute communication and activation, receptively. SRIA and PMA denote semantic-related instance attention, and patch mixing and activation, receptively. }\vspace{-3mm}
	\label{Tab:ablations}
	\begin{tabular}{c|cc|cc|ccc|ccc|ccc}
		\hline
		\multicolumn{1}{c|}{\multirow{2}{*}{baseline}} & \multicolumn{2}{c|}{IMSE}                            & \multicolumn{2}{c|}{SMID}       & \multicolumn{3}{c|}{CUB} & \multicolumn{3}{c|}{SUN}& \multicolumn{3}{c}{AwA2} \\ \cline{2-14} 
		\multicolumn{1}{c|}{}                          & \multicolumn{1}{c|}{IASA} & \multicolumn{1}{c|}{ACA} & \multicolumn{1}{c|}{SRIA} & PMA & $U$      & $S$      & $H$      & $U$      & $S$      & $H$  & $U$      & $S$      & $H$     \\ \hline \hline
		\checkmark  & \multicolumn{1}{c|}{}   &   &\multicolumn{1}{c|}{}       &     & 59.8  & 68.4  & 63.8 &43.8 &30.6&36.0 & 58.0  & 81.6  & 67.8  \\
		\checkmark  & \multicolumn{1}{c|}{}   &   &\multicolumn{1}{c|}{\checkmark}       &     & 63.5  & 71.11  & 67.1 &57.7 &32.2 &41.3  & 63.2  & 75.6  & 69.3  \\
		\checkmark & \multicolumn{1}{c|}{} &  &\multicolumn{1}{c|}{ \checkmark}& \checkmark & 70.0   & 70.0  & 70.0  &60.3 &41.8 &49.4 & 65.0 & 77.3   & 70.6   \\
		\checkmark  & \multicolumn{1}{c|}{ \checkmark}  &   & \multicolumn{1}{c|}{ \checkmark}  & \checkmark   & 70.0     & 72.8   & 71.3  &61.4  &43.9 & 51.2  & 71.1   & 78.1   & 74.5   \\
		\checkmark  & \multicolumn{1}{c|}{ \checkmark} & \checkmark  & \multicolumn{1}{c|}{ \checkmark} & \checkmark   & 70.1   & 77.8   & 73.8 &61.7 &45.4 &52.3  & 73.6   & 77.3   & 75.4  \\ \hline
	\end{tabular}
\end{table*}
 To give a clear insight into each component in our framework, we perform ablations to analyze the effectiveness of significant components. \cref{Tab:ablations} summarizes the results of ablation studies. 
Firstly, the baseline method means that we directly compute the scores between the visual features extracted from ViT and the category prototypes to infer the category.
Compared to baseline, the model only using SRIA which directly applies the sharing attributes to conduct semantic-related instance adaption achieves a significant improvement.
We then add the PMA in SMID, and the $H$ metric further increases, verifying that spatial exploration captures discriminative features that promote category inference.
After that, we incorporate IASA into this model to learn the instance-motivated semantic attribute.
Therefore, the model can get the $H$ improvements of 1.3\%, 1.8\%, and 3.9\% on CUB, SUN, and AwA2, respectively, benefiting from semantic-visual mutual adaptation.
In addition, when ACA module is conducted in IMSE to form our full PSVMA, the model realizes performance increases on both CUB and AwA2 datasets. We think that such improvement stems from the compacted and activated attributes by ACA.
By combining all the components, our full model progressively adapts visual and semantic representations in a mutual reinforcement manner, achieving $H$ improvements of 10.0\%, 16.3\%, and 7.6\% on CUB, SUN, and AwA2 over the baseline, respectively.

\subsection{Hyperparameter Analysis} 
 \begin{figure*}[t]
	\centering
	\includegraphics[width=1.0\linewidth]{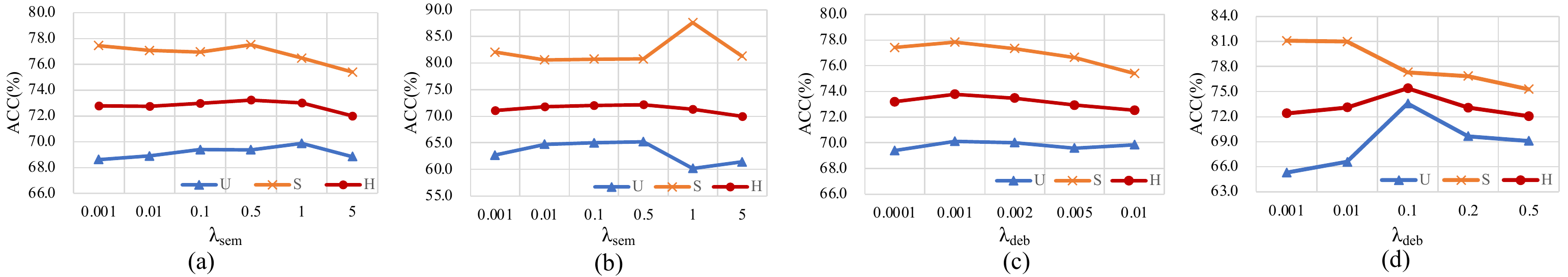}\vspace{-2mm}
	\caption{Effect of loss weights. $\lambda_{sem}$ on (a) CUB and (b) AwA2. $\lambda_{deb}$ on (c) CUB and (d) AwA2. }\vspace{-4mm}
	\label{fig:loss}
\end{figure*}
\begin{figure}[t]
	\centering
	\includegraphics[width=0.9\linewidth]{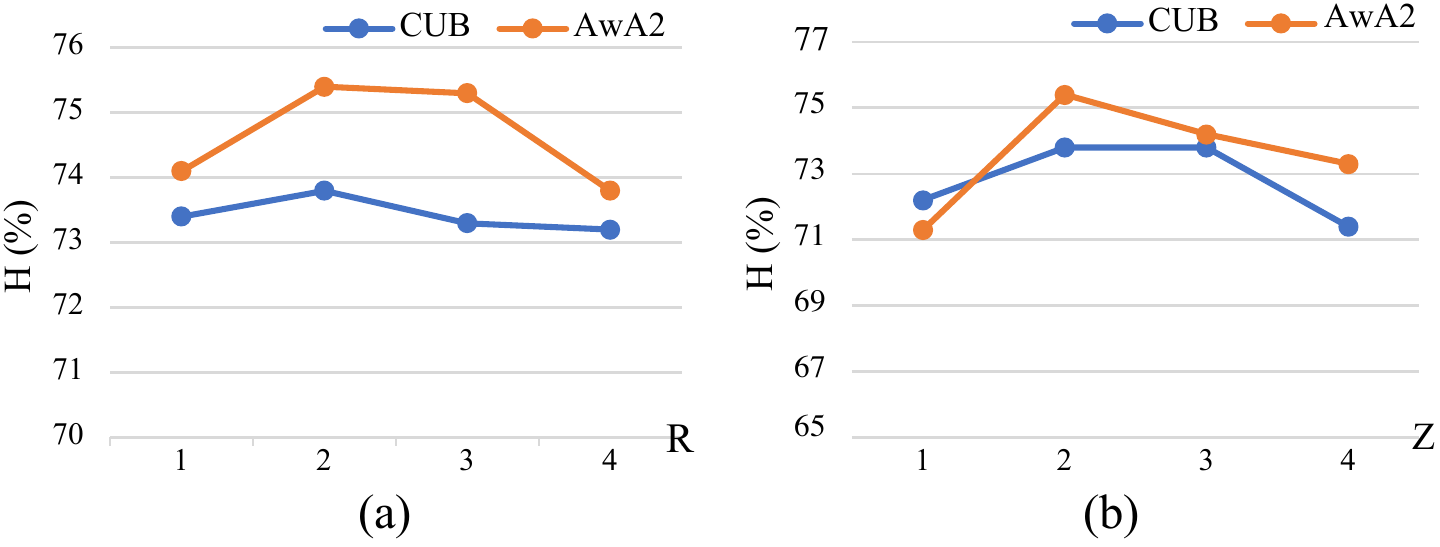}\vspace{-2mm}
	\caption{Effect of (a) $R$, (b) $Z$ on CUB and AwA2 datasets.}\vspace{-5mm}
	\label{fig:kth}
\end{figure}
\noindent
\textbf{Effect of $\lambda _{sem}$ and $\lambda _{deb}$ in Loss.}
In \cref{fig:loss}, we evaluate the effect of loss weights $\lambda _{sem}$ and $\lambda _{deb}$ in \cref{eq:loss}. We first set the value of $\lambda _{deb}$ to $0$ to analyze the effect of hyper-parameter $\lambda _{sem}$. As $\lambda _{sem}$ raises, the harmonic mean rises slowly at first and then decreases when $\lambda _{sem} >0.5$.
Large $\lambda _{sem}$ over emphasizes the knowledge on the seen domain by \cref{eq:L_sem}, resulting in poor generalization capability.
Thus, we set $\lambda _{sem}=0.5$ for CUB and AwA2.
Then, we gradually increase the value of $\lambda _{deb}$. When more attention has been paid to pursuing the distribution consistency between seen and unseen predictions, we get better unseen performance. However, both $\lambda _{sem}$ and $\lambda _{deb}$ can not be too large to avoid squeezing the capacity of classification loss, resulting in identification accuracy reduction. 
Therefore, we fix $\lambda _{deb}$ to $0.001$ for CUB and $0.1$ for AwA2 in our experiments.

\begin{figure}[t]
	\centering
	\includegraphics[width=0.8\linewidth]{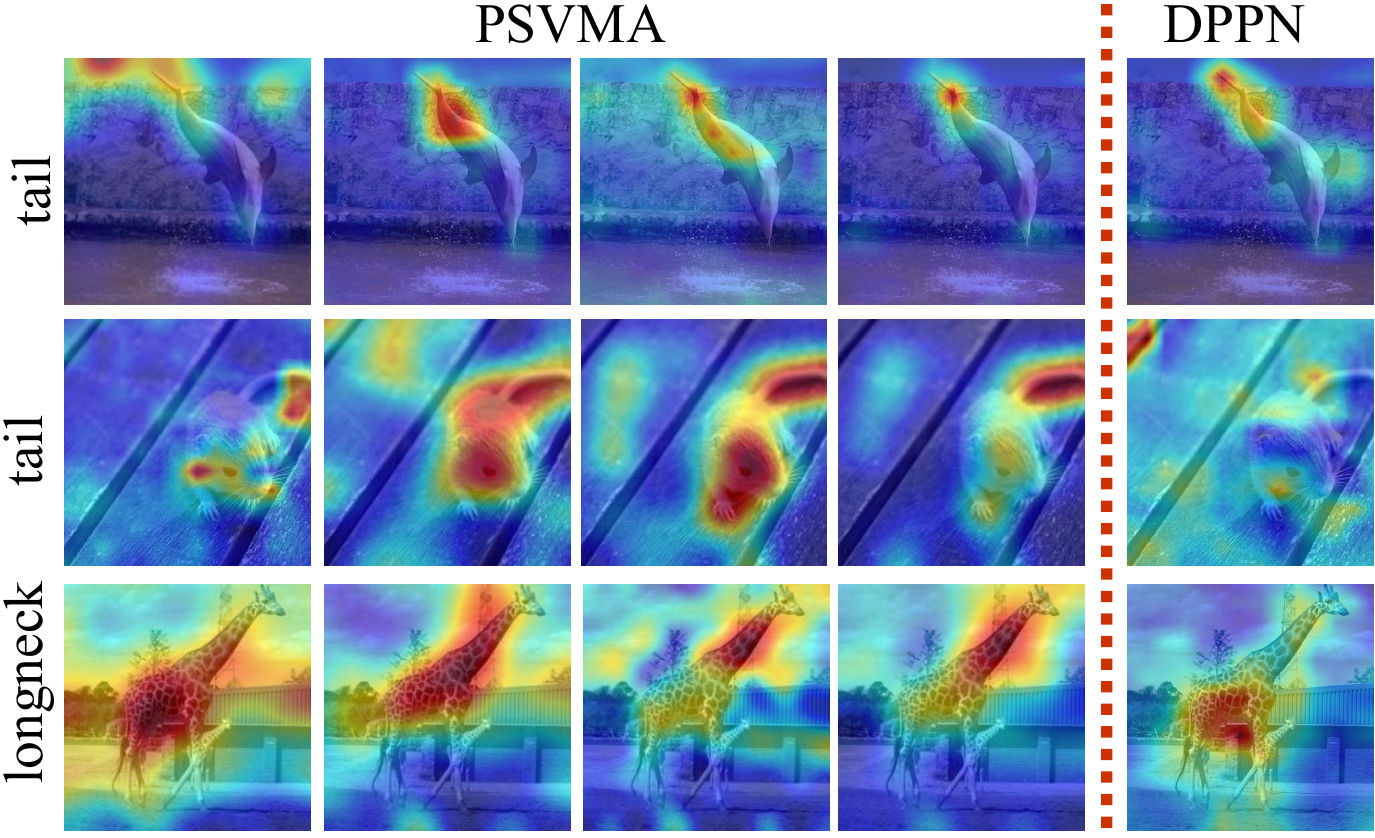}\vspace{-3mm}
	\caption{Visualization of attention maps of our PSVMA and  DPPN\cite{DPPN2021}. The 1-4 columns imply the effectiveness of progressive learning.} \vspace{-7mm}
	\label{fig:attention}
\end{figure}

\noindent
\textbf{Effect of $R$, $Z$ in PSVMA.}
To achieve progressive semantic-visual mutual adaption, 
PSVMA deploys $Z$ DSVTMs with $R$ recurrent IMSEs between different visual layers and semantic attributes. 
As shown in \cref{fig:kth} (a) and (b), when $R=Z=2$, the model gains the best $H = 73.8\%$ and $75.4\%$ on CUB and AwA2 datasets.
This demonstrates that the progressive adaption for instance-centric attributes and unambiguous visual representations are beneficial for semantic-visual interactions, improving the transferability for GZSL. 
However, $H$ decreases when $R$ and $Z$ are larger than 2. 
This is due to the excessive learning of instance-related information adapted over four times, which limits the classification performance. 
When the progressive learning exceeds 4 adaption, the model tends to learn the instance-related information of the seen domain, which affects its knowledge transfer ability to the unseen domain, thus leading to the performance drop.
Therefore, we choose $R=Z=2$ as default settings for progressive learning.
Furthermore, \cref{fig:attention} intuitively demonstrates the effectiveness of our progressive adaption.
With 4-times learning (column 1-4), the attribute localization gets more precise. The same attribute ``tail''  for distinct images gets more specialized (row 1-2).  Besides, our localization is much more accurate compared with DPPN\cite{DPPN2021}.

\section{Conclusion}
In this paper, we aim to semantic disambiguation and propose a progressive semantic-visual mutual adaption (PSVMA) network by deploying the dual semantic-visual transformer module (DSVTM) executed among different visual layers and attribute prototypes. 
Specifically, DSVTM adapts the sharing attributes to different input images and acquires instance-centric attributes, enabling the recast of semantic-visual pair.
With the matched pair, DSVTM constructs accurate cross-domain interactions and distills unambiguous visual representations adapted to target semantics, improving the transferability. 
 Besides, a debiasing loss mitigates seen-unseen bias to assist the knowledge transfer process for GZSL.
 Extensive experiments on three public datasets show the superiority of our PSVMA. The codes using MindSpore \cite{mindspore} will also be released at {\small{https://gitee.com/chunjie-zhang/psvma-cvpr2023}}. 

 \vspace*{2mm}\noindent {\bf Acknowledgements:} This work was supported in part by National Key R\&D Program of China (No.2021ZD0112100), National Natural Science Foundation of China (No.61972023, 62072026, U1936212), Beijing Natural Science Foundation (L223022, JQ20022), and the Open Research Fund of The Laboratory of Cognition and Decision Intelligence for Complex Systems, Institute of Automation, Chinese Academy of Sciences. We gratefully acknowledge the support of MindSpore, CANN (Compute Architecture for Neural Networks) and the Ascend AI Processor used for this research. 

{\small
\bibliographystyle{ieee_fullname}
\bibliography{egbib}
}

\end{document}